\documentclass{article}

\usepackage[preprint,nonatbib]{cml4impact_2022}

\usepackage[utf8]{inputenc} 
\usepackage[T1]{fontenc}    
\usepackage{hyperref}       
\hypersetup{hidelinks}
\usepackage{url}            
\usepackage{booktabs}       
\usepackage{amsfonts}       
\usepackage{nicefrac}       
\usepackage{microtype}      
\usepackage{xcolor}         
\usepackage[backend=biber,style=vancouver,]{biblatex}

\usepackage{amsmath}
\usepackage{mathtools}

\definecolor{LightGray}{gray}{0.95}

\title{Out-of-sample scoring and automatic selection of causal estimators}
\author{%
  Egor Kraev \\
  Wise Plc\\
  \texttt{egor.kraev@wise.com} \\
  \And
  Timo Flesch \\
  OSG Digital\\
  University of Oxford \\
  \texttt{timo.flesch@psy.ox.ac.uk} \\
  \And
  Hudson Taylor Lekunze \\
  University of Tartu \\
  \texttt{hudson.lekunze@outlook.com} \\
  \And
  Mark Harley \\
  Wise Plc\\
  \texttt{mark.harley@wise.com} \\
  \And
  Pere Planell Morell \\
  Wise Plc\\
  \texttt{pere.planell@wise.com}
}

\begin{document}

\maketitle

\begin{abstract}
Recently, many causal estimators for Conditional Average Treatment Effect (CATE) and instrumental variable (IV) problems have been published and open sourced, allowing to estimate granular impact of both randomized treatments (such as A/B tests) and of user choices on the outcomes of interest. However, the practical application of such models has been hampered by the lack of a valid way to score the performance of such models out of sample, in order to select the best one for a given application.
We address that gap by proposing novel scoring approaches for both the CATE case and an important subset of instrumental variable problems, namely those where the instrumental variable is customer access to a product feature, and the treatment is the customer's choice to use that feature.

Being able to score model performance out of sample allows us to apply hyperparameter optimization methods to causal model selection and tuning. We implement that in an open source package that relies on DoWhy and EconML libraries for implementation of causal inference models (and also includes a Transformed Outcome model implementation), and on FLAML for hyperparameter optimization and for component models used in the causal models.

We demonstrate on synthetic data that optimizing the proposed scores is a reliable method for choosing the model and its hyperparameter values, whose estimates are close to the true impact, in the randomized CATE and IV cases. Further, we provide examples of applying these methods to real customer data from Wise.
\end{abstract}

\section{Introduction}
Measuring impacts of actions or choices, conditional on customer characteristics, is a crucial problem for any business. This can be used for optimizing these choices, such as which layout to show or which promotion to send to a particular user, or for measuring the impact of the choices we are less able to directly control or randomize, for example of suspending a user who has been flagged as suspicious, or of the user choosing to use a particular product feature.

Causal inference models can help with that; yet an as yet unsolved challenge is how, in a given application, to choose a particular model among the many options available, and how to tune its parameters. In automated selection and tuning of supervised machine learning models, also known as automated machine learning (AutoML), the key concept is out-of-sample scoring, that is comparing the models' performance on a validation dataset that wasn't used in training. Here, performance is defined by a distance metric comparing model predictions and target values. The model (including hyperparameter choices) that performs best on the validation set is selected by the algorithm. 

We extend this approach to causal inference models. This is not a trivial task as these models estimate the impact of an action, which is not directly observable (we can only observe an outcome of e.g. sending an email to a customer, but we can't \emph{not} send the same email to the same customer at the same time to observe the \emph{impact} of sending the email). This is known as The Fundamental Problem of Causal Inference \cite{pearl2009causality}.

\begin{figure}
  \centering
  \includegraphics[scale=0.165]{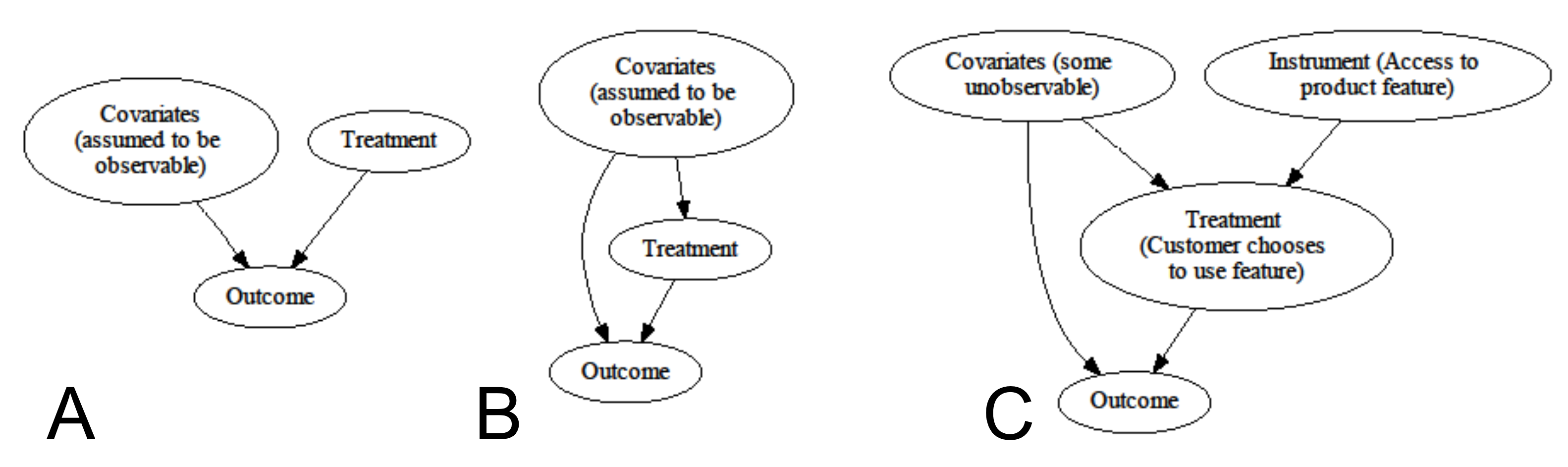}
  \caption{\textbf{Types of causal graphs we consider}}
  \label{fig:causal_graphs}
\end{figure}

We consider three classes of practically important problems: firstly, randomized CATE, where the treatment assignment is fully random, such as may be the case in an A/B test (Figure~\ref{fig:causal_graphs}A). The second problem is CATE with known confounders (Figure~\ref{fig:causal_graphs}B); this can arise for example when we assign treatment with a degree of randomness, but biased according to prior opinions on which kind of customer responds best to treatment). Finally, we consider the special subset of instrumental variable models where access to a product feature is the instrumental variable, and the customer's decision to use it is the treatment  (Figure~\ref{fig:causal_graphs}C).

 We introduce new ways of scoring model performance, in particular out of sample, that are specific to the causal inference problems in question - one score (energy distance) that is valid for Conditional Average Treatment Effect (CATE) models under unconfoundedness and for instrumental variable models, and another (Normalized ERUPT) that is valid for CATE problems also with observed confounders. We test them on synthetic data and then apply them to Wise customer behavior data.
 
We implement our approach as an open-source pip-installable Python package \texttt{AutoCausality}, which capitalises on Microsoft's \texttt{FLAML} package for efficient hyperparameter tuning \cite{wang_flaml_2021} and \texttt{DoWhy} and \texttt{EconML} packages for causal model estimation \cite{keith_battocchi_econml_2019, sharma_dowhy_2020} \footnote{\hyperlink{https://github.com/transferwise/auto-causality}{https://github.com/transferwise/auto-causality}}.
 
For clarity, we restrict our discussion to the case of a binary treatment; however the approach works equally well for multi-valued or continuous treatments\footnote{The package currently only supports categorical treatments, including multi-valued ones. However, the scoring methods we use, in particular energy distance, are valid for continuous treatments as well, and many of the underlying \texttt{EconML} estimators also support these.}. 

\section{Scoring methods}
\subsection{Existing methods}
Perhaps the most popular score in the literature for CATE problems is the Qini coefficient \cite{radcliffe_using_2007}. Its main limitations are firstly, that its derivation is only applicable to randomized CATE scenarios, and secondly, that it doesn't readily generalize beyond binary treatments.

Another existing method is the R-scorer \cite{schuler_comparison_2018} as implemented  for example in \texttt{EconML}. It relies on fitting component models for treatment and outcome as function of confounders, which firstly has high computational cost (especially if, as we do, one defaults to AutoML for component model construction), and secondly creates a seemingly paradoxical situation of effectively using a model to score another model, leaving open the question of why the output of the former model is to be considered more trustworthy than that of the latter. 

A final popular family of scores is known as policy value, or ERUPT for Estimated Response to Proposed Treatments (\cite{hitsch_heterogeneous_2018}, Equation (3)). This takes an arbitrary treatment assignment (`policy'), specified as a function of customer features, and the results of an existing experiment (which could use any other treatment assignment including strictly random, as long as the propensity to treat is strictly positive), and combines them to produce an unbiased estimate of expected outcome had we applied that policy.

It does that by taking all data points where the desired policy coincides with the one actually used in the experiment (we are guaranteed to find such given a large enough sample size due to positivity of the propensity function), and scales the outcomes of these cases with the inverse of the propensity function to obtain an unbiased estimate (proof in \cite{hitsch_heterogeneous_2018}). This is closely related to importance sampling \cite{10.2307/1913641}.

The advantage of this metric is that it has the same units as the outcome (such as monetary units or number of conversions), which makes it easier to interpret in a business context. However, both its strength and its weakness is that it requires us to specify a policy to be used for calculating the score. This is a strength as it allows us to plug in business-relevant metrics and be explicit about treatment cost (for example, `treat if expected impact exceeds treatment-specific cost'), but a weakness as it requires us to make a choice of policy, which is at odds with our desire to automate model selection as much as possible.
\subsection{Normalized ERUPT for CATE models}
For CATE models, both and without confounding, we use a policy value score chosen to optimize the prediction of impact heterogeneity. Let us designate by $i_j$ the impact of (a binary) treatment on unit $j$, and $\bar i$ is its average value. The naive approach would be to use (and let the fitter  maximize) the policy "\texttt{treat iff $i_j> 0$}". However, if the impact is heterogenous but positive throughout, this score will teach us little as it will be identical for all broadly correct models.

Instead, we introduce Normalized ERUPT as the policy value of the policy "\texttt{treat iff $i_j > \bar i$}". Thus while each individual model, when fitting, tries to estimate the pointwise conditional impact itself, by using this score for subsequent out-of-sample comparison of models we select for their ability to estimate \emph{deviations} from the average impact, with the expectation that the average impact estimate will also have to be accurate for this to work.
      
\subsection{Energy distance}  
This is applicable, with slight variations, both to randomized CATE models and to IV models. 

Let us first consider the IV case: suppose we have a set of customers $X$, where each element $x\in X$ is a feature vector for a single customer. We randomly split them into the control set $X_0$ of customers who don't get access to a particular new product feature, and the set $X_1$ of customers who get access. Of the latter, the subset $X_{1, f}$ of customers choose to use the product feature, and the rest don't. We further observe some outcome $y$, such as revenue in the 30 following days, for all customers in $X$. 
If we want to measure the impact of the users using the product feature, we can begin by assuming that the product feature only impacts the outcome for those users who actually used it. We can then treat access to it as an instrumental variable, and usage as the treatment.

How can we compare the performance of different IV models out of sample?

We propose the following approach: let's call the true impact of using the product feature $dy$, which by definition may only be nonzero for the customers in $X_{1, f}$. As we suppose the only impact of making the product feature available is through its usage, the conditional distribution of the corrected outcome $\hat y \coloneqq y - dy$ given the customer features, $P\left(\hat y | x\right)$ will be identical between the sets $X_0$ and $X_1$.

This allows us to compare IV models out of sample, by holding out a fraction of both $X_0$ and $X_1$ as a validation set (let's call them $X^v_0$ and $X^v_1$), training the IV models on the remaining data, and using each of the models to calculate $\hat y_m$ on $X^v_1$ (the index $m$ goes over the different models). 

We can then introduce some measure of distributional similarity $d$, and use it to compute the similarity between the distributions of $\hat y^m$ over $X^v_0$ and $X^v_1$, and use that as a score for model $m$, with higher values denoting worse models, so 
$$ S_m = d( P\left(\hat y_m | x\right) ~\text{over}~ X^v_0, P\left(\hat y_m | x\right) ~\text{over}~ X^v_1) $$


As the customers' access was assigned randomly, the distribution of customer features will be identical in $X^v_0$ and $X^v_1$. Therefore, if  we combine the feature vector and the corrected outcome into a single extended feature vector, a `perfect' IV model will result in the extended feature vectors' distributions also being identical. 

The problem then becomes to define a distance between the two distributions from which the respective sets of extended feature vectors have been sampled (no longer conditional distributions).

The approach we use is energy distance \cite{SZEKELY20131249}, as implemented e.g. in the \href{https://dcor.readthedocs.io/en/latest/theory.html#energy-distance}{\texttt{dcor} package}.

\section{Using scoring methods for automatic model selection and tuning}
The search happens as follows: the API expects a list of the applicable estimators that should be searched over (or chooses the default set) for one of the problems above, as well as the dataset to be used for fitting. We specify a hyperparameter search space for each of the relevant causal inference models, then combine them into a hierarchical search space, first choosing a model, then a set of hyperparameters for it. Note that the search spaces for the \texttt{FLAML} component models used as part of a given causal model (if any), are not part of this search space, they are encapsulated in the component models. 

The supplied dataset is split into train and validation (the test set is set aside earlier), these are stored in the \texttt{AutoCausality} class instance and are the same for all fits of that optimization run. 

We first fit each of the chosen estimators using a default hyperparameter set for it (taken from \texttt{EconML} docs) and calculate the resulting score on the validation set; then let the \texttt{FLAML} tuner take the results of these runs and search over the composite search space (models/hyperparameters) to find the causal inference model with the best score, within the allocated time budget. 

Most causal inference models use supervised ML models as components. The most important of these is the `propensity function', that is the a priori probability that a given unit will be treated (e.g. that a customer will be chosen to be sent an email), given its features. In all causal models we are aware of, this is treated as a classifier, and its fitting a supervised ML problem. We also follow that approach, defaulting to \texttt{sklearn}'s \verb|DummyClassifier(strategy='prior')| for the cases when we know the assignment to be random, and \texttt{FLAML}'s \verb|AutoML(task='classification')| for all other cases. However, we also allow any user-supplied classifier model that supports the \texttt{sklearn} \verb|fit/predict_proba| API.

Most advanced causal inference models also require other regression and/or classifier component models to be supplied, to represent certain relationships in the causal graph at hand. For now, we use \verb|DummyClassifier| when we know the relationship in question to be random, and \texttt{FLAML} regressor/classifier in all other cases. This is not a fundamental limitation of our approach, but merely a convenience choice to outsource component model selection. 

Note that inside each estimator fit, if the model in question includes a \texttt{FLAML} component classifier or regressor, we do a hyperparameter search over classifier/regressor models to fit that classifier/regressor; we thus use \texttt{FLAML} twice in a nested fashion, once to fit component models, and once for the high-level search over causal estimators and their hyperparameters.

\section{Performance on synthetic data}
\begin{figure}
  \centering
  \includegraphics[scale=0.55]{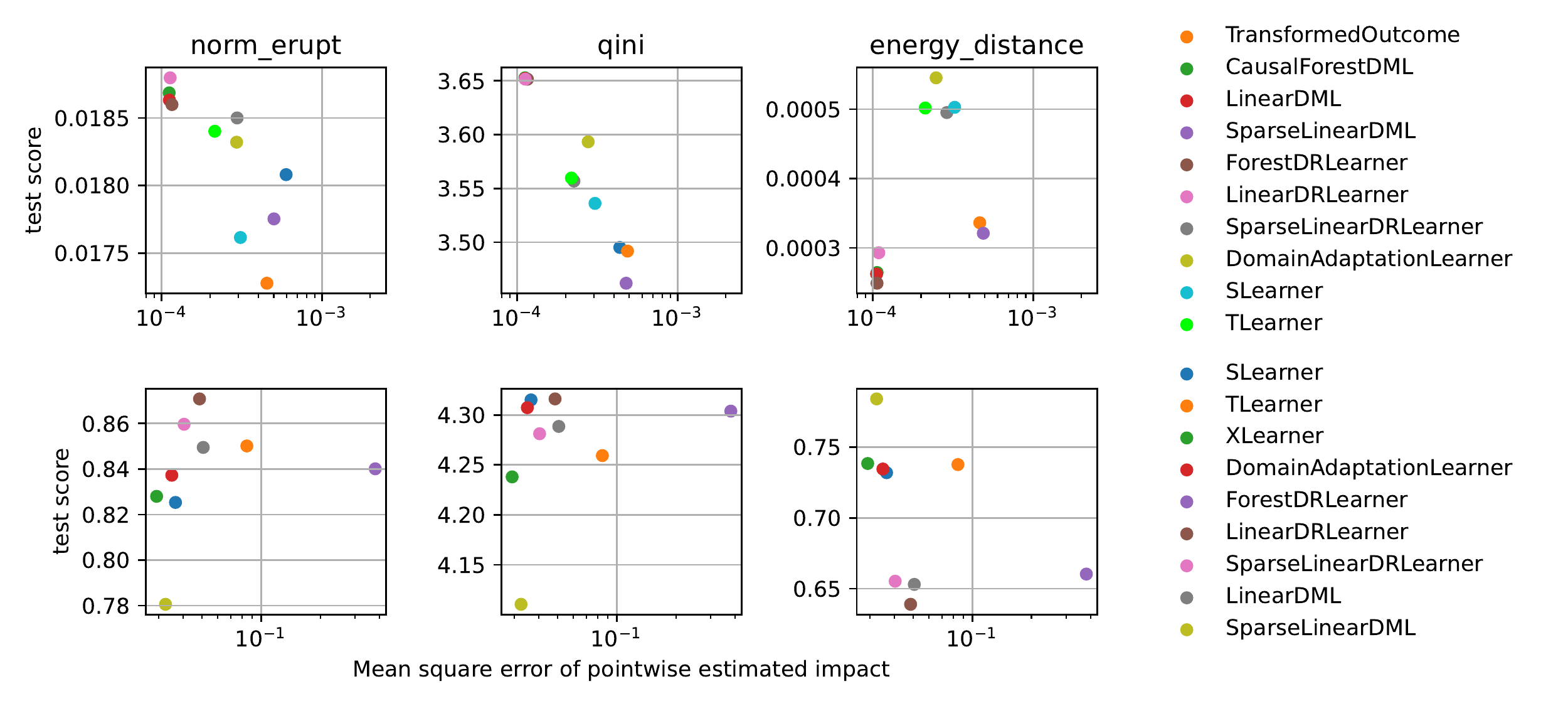}
  \caption{\textbf{Scoring CATE estimators on synthetic data. Top row: Randomized treatment assignment, Bottom row: fully observable confounders.} Score vs MSE, for different scores. For normalized ERUPT and Qini high values are better, while for energy score low values are better. } 
  \label{fig:scores_cate_rct}
\end{figure}

To verify the effectiveness of the proposed scores as proxies for the quality of a given model's estimate of the causal impact, we run the above optimization on synthetic data, to directly observe the relationship. The details of the construction of the synthetic dataset are given in Appendix~\ref{appendix_synth}.
\paragraph{Randomized CATE}
In this case, corresponding to a traditional A/B testing scenario of Figure~\ref{fig:causal_graphs}A, we conduct three optimization runs, with normalized ERUPT, Qini score, and energy score, respectively,and plot the respective score of each run against the mean square error (MSE) of its impact estimate in the top row of Figure~\ref{fig:scores_cate_rct}. 
We see that each of the three scores provides a reliable guidance for choosing the model with the best MSE for the synthetic dataset at hand. Figure~\ref{fig:mse_rct} illustrates that the model chosen using the energy distance score shows the best performance.

\paragraph{CATE with confounding}
The results for CATE with known confounders, corresponding to Figure \ref{fig:causal_graphs}B, are shown in the bottom row of Figure~\ref{fig:scores_cate_rct}. Note that neither the Qini score, nor the above derivation of the energy score are strictly speaking applicable to this case, but we wanted to assess whether they would still perform well in practice. Unfortunately, in this case none of the scores perform that well. However for none of the scores did the best-scoring model have one of the worst MSEs, suggesting that even in this case the suggested scores can be useful for rough model selection, even if not for model fine-tuning. 

\begin{figure}
  \centering
  \includegraphics[scale=0.5]{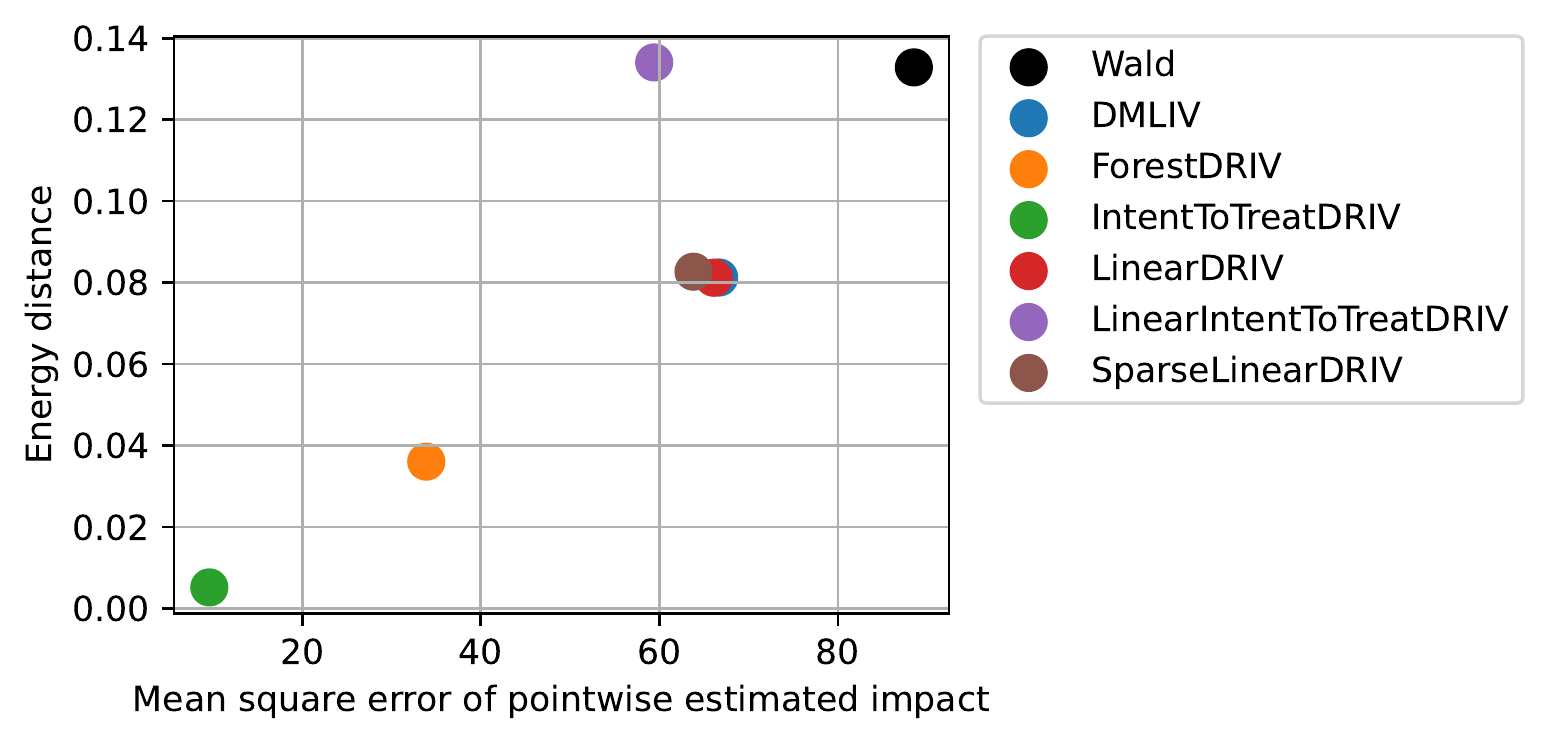}
  \caption{\textbf{Scoring IV Estimators on synthetic data.}}
  \label{fig:IV_synth}
\end{figure}

\paragraph{Instrumental variables}
Figure \ref{fig:IV_synth} shows the results for the instrumental variable case, corresponding to Figure  \ref{fig:causal_graphs}C. There we compare the performance of the listed estimators using the Wald Estimator\cite{wald_1940} as a baseline. We see that for this synthetic dataset, energy distance provides a reliable way of choosing the most accurate estimator. 

\section{Performance on real data}
The above methods are already seeing application across Wise. This section presents two examples. The first is assessing the impact of making a new product feature (Auto-conversion) available. We can assess that in two ways: firstly, by measuring the impact of making the feature available (known as Intent To Treat, or ITT), which we can randomize. Secondly, we can use access to the feature as an instrumental variable to estimate the impact of the users choosing to use the feature. 

The second case is one where randomization is not possible, namely the impact of suspending customers the likelihood of the customer doing a transaction once reactivated. 

As we don't know the true impact when working on real data, to evaluate the quality of the fit we compare the model scores firstly, to a baseline model (naive propensity-score weighted averages with 1\% multiplicative random noise in the CATE case, Wald estimator in the IV case), and secondly, across validation and test sets. If the scores across validation and test set correlate strongly, and systematically outperform the baseline model, we conclude that the fit was successful. 

\begin{figure}
  \centering
  \includegraphics[scale=0.4]{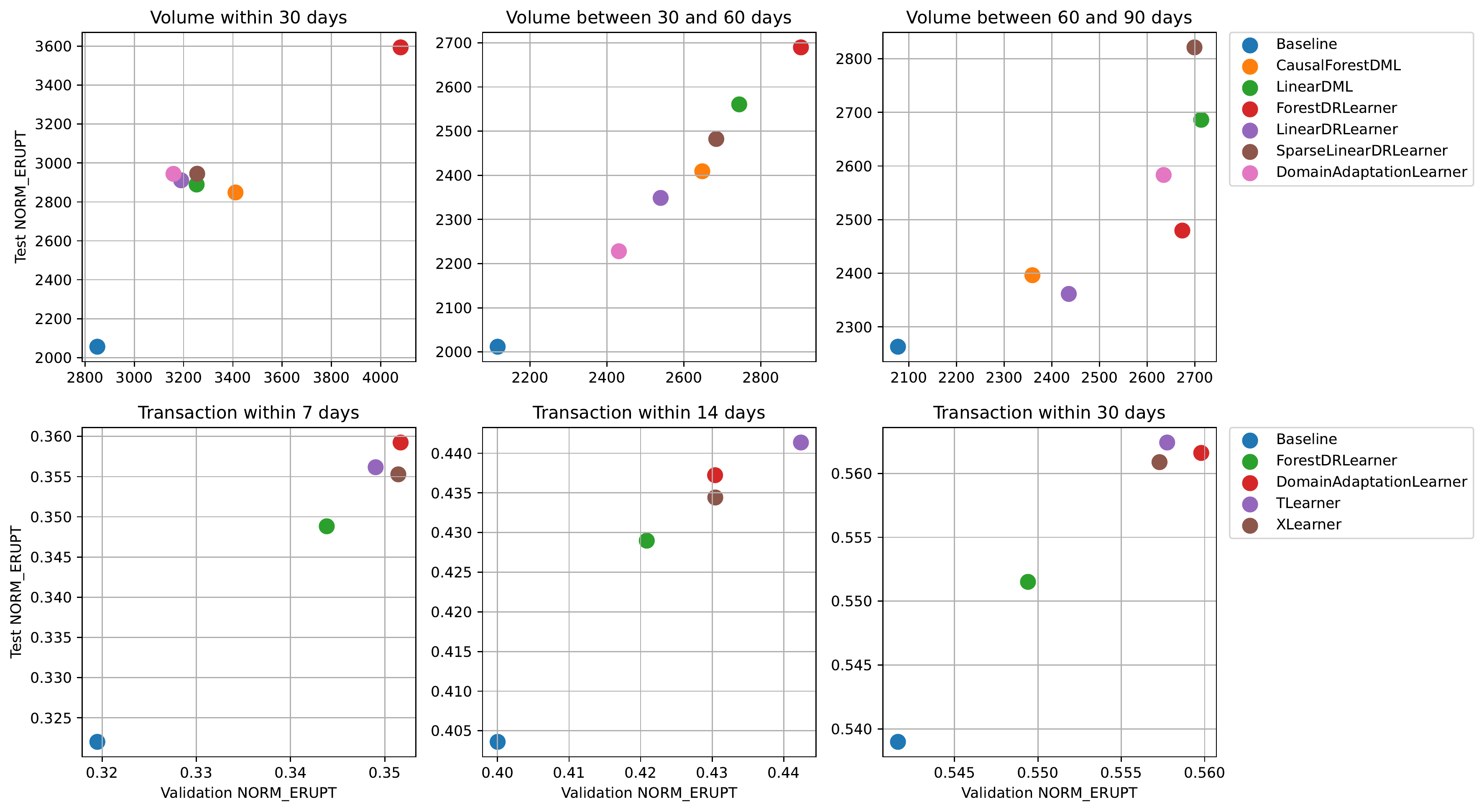}
  \caption{{\bf Score consistency between test and validation. } {\bf Top:} impact of (randomized) access to product feature on volume. {\bf Bottom:} Impact of suspending customers on their transacting again 
  after reactivation}
  \label{fig:autoconversion2}
\end{figure}

The top row of Figure~\ref{fig:autoconversion2} presents such results for the ITT analysis on customer volumes over different time horizons after treatment. We see that the scores on test and validation sets are indeed strongly correlated, and systematically outperform the baseline. For reasons of space, we only show the results for the normalized ERUPT metric. 

The IV analysis is ongoing. The main challenge is that with a low conversion rate, we need quite a large sample size to get reliable results; yet the IV causal estimators are particularly demanding in terms of RAM required to fit the estimator.

The final analysis is that of the impact of user suspensions (bottom row of Figure~\ref{fig:autoconversion2}), interesting as no random assignment is possible in this case. This also shows a strong correlation between validation and test scores. 

In both cases, it's interesting that no estimator consistently outperforms all the others, thus demonstrating the value of automated estimator selection.

\section{Conclusion}
Causal inference is an exciting domain that holds great promise for business impact even in the seemingly mature areas such as analysis of A/B test results. One of the obstacles to its wider adoption is currently the variety of available estimators, with no clear criteria to choose between them.

Scoring causal estimators out of sample allows us to extend the power of AutoML for estimator selection and hyperparameter tuning, to three practically important classes of causal inference problems. This will make these methods easier to use for the broad audience that they deserve. 

\printbibliography
\appendix
\section{Construction of synthetic datasets}

\subsection{CATE dataset without confounders}
\label{appendix_synth}
To simulate a randomized control trial (RTC) corresponding to the graph shown in figure  \ref{fig:causal_graphs}A, we used the following data generating process (DGP):  \\
Let $X^{Nxd}$ be the matrix of $N$ observations and $d$ covariates, $T^{nx1}$ the vector of treatment assignments and $Y^{nx1}$ the vector of outcomes. 
We make the following assumptions: 
\begin{itemize}
\item binary treatments 
\item fully random propensity to treat (unconfoundedness) 
\item five continuous, normally distributed covariates 
\item no interaction between treatment effects and covariates 
\item independence of the covariates, i.e. $\Sigma = \sigma^2I$
\item no additive noise in the outcomes, i.e. $\epsilon=0$ 
\end{itemize}

Then, the data is generated according to the following equations:
\begin{align*}
& X_i \sim \mathcal{N}(0,\Sigma) \\
& T_i \sim Bernoulli(0.5) \\
& Y_i = \tau(X_i) T_i + \mu_0(X_i) + \epsilon
\end{align*}
where $i$ indexes individual units, $\tau$ describes the following true treatment effect, which depends linearly on all covariates:
\begin{equation*}
\tau(X_i) = X_ib^T + e
\end{equation*}
where $b$ is a 1xd vector of $b_i \sim U(0.4,0.7)$ weights for each covariate and $e \sim \mathcal{N}(0,0.05)$ Gaussian noise.  
... and  $\mu_0(x)$ describes the following transformation of the covariates (to keep things interesting):
\begin{equation*}
\mu_0(X_i) = X_{i,1} \otimes X_{i,2} + X_{i,3} + X_{i,4} \otimes X_{i,5} 
\end{equation*}

\subsection{CATE dataset with confounders}
To simulate a dataset from an observational study, i.e. with confounders, corresponding to graph shown in figure  \ref{fig:causal_graphs}B, we used the same DPG as shown above for the RCT data, but this time with treatment assignments that were dependent on a subset of covariates as follows:
\begin{equation*}
T_i \sim Bernoulli \left( clip\left(\frac{1}{1+exp(X_{i,1} \otimes X_{i,2} + 3*X_{i,3}}, 0.1, 0.9 \right)\right)
\end{equation*}

\subsection{Instrumental Variables}
The dataset simulating an observational study with an instrumental variable that determined who would be offered a treatment, corresponding to the graph shown in figure \ref{fig:causal_graphs}C, was generated in a similar fashion as the one above, but treatment allocation $T$ depended on the instrument $Z$ as follows:

\begin{align}
W \sim \; & \text{Normal}(0,\, I_{n_w})   \tag{Observed confounders}\\
Z \sim \; & \text{Bernoulli}(p=0.5)   \tag{Instrument}\\
\nu \sim \; & \text{U}[0, 5] \tag{Unobserved confounder}\\
C \sim \; & \text{Bernoulli}(p=0.8 \cdot \text{Sigmoid}(0.4 \cdot X[0] + \nu)) \\
C0 \sim \; & \text{Bernoulli}(p=0.006)  \\
\theta = & \; 7.5\cdot X[2]\cdot X[8] \tag{Nonlinear effect}\\
T = \; & C \cdot Z + C0 \cdot (1-Z)  \tag{Treatment}\\
y \sim \; & \theta \cdot T + 2 \cdot \nu + 5 \cdot (X[3]>0) + 0.1 \cdot \text{U}[0, 1]  \tag{Outcome}
\end{align}

\section{Additional plots}
\begin{figure}
  \centering
  \includegraphics[scale=0.7]{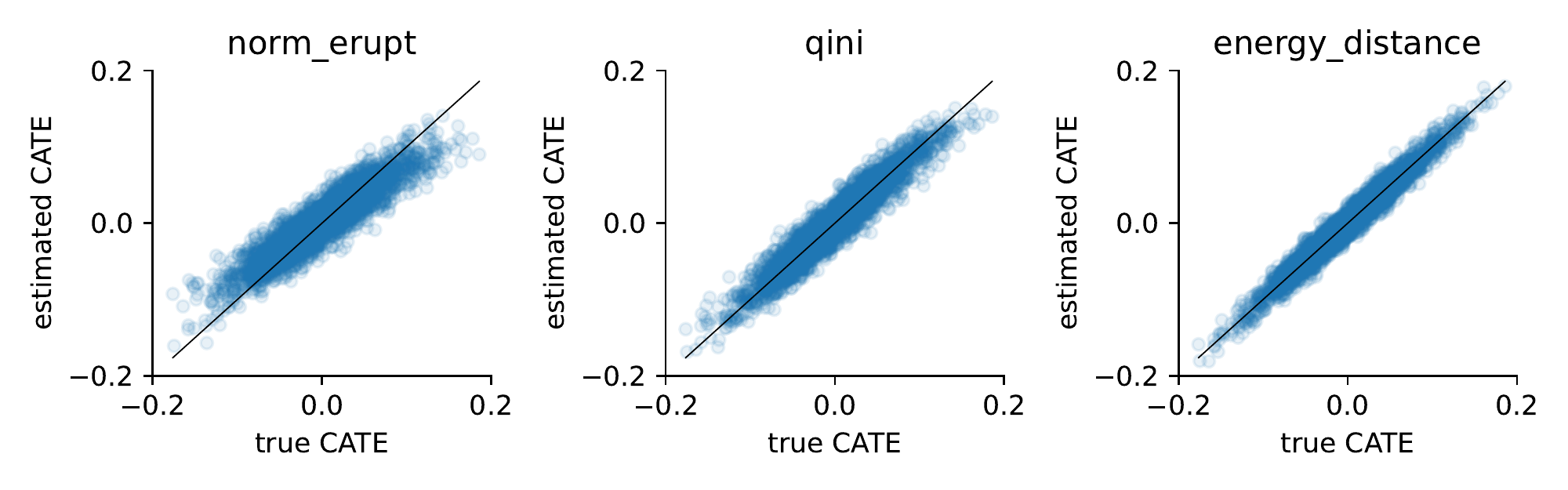}
  \caption{Randomized treatment, synthetic data. MSE between ground truth and estimated CATE of best-fitting model on the test set.}
  \label{fig:mse_rct}
\end{figure}

Here we provide a couple of plots to give some additional colour to the results. Figure~\ref{fig:mse_rct} plots pointwise effect estimates for the unconfounded case against the true effect values in the synthetic data, for the best model according to each score. We see that the energy distance makes for the best selection criterion in this case.

\begin{figure}
  \centering
  \includegraphics[scale=0.45]{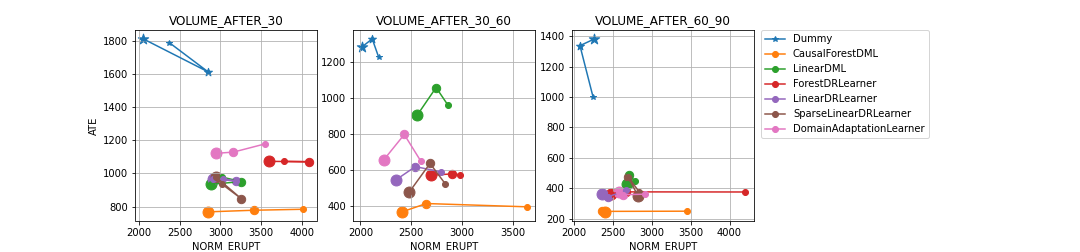}
  \caption{AutoConversion feature: scores vs ATE estimates. Color denotes different estimators, smallest dot represents the values on the training set, medium sized dot those on the validation set, and the largest one those on the test set.}
  \label{fig:autoconversion1}
\end{figure}

Figure~\ref{fig:autoconversion1} plots the estimated average effect (ATE) against the score for each of the estimators for the ITT on real data usecase. We see that although the treatment assignment (access to product feature) was randomized, the naive estimator (labeled as "Dummy") significantly over-estimates the impact, but also that different estimators with comparable Normalized ERUPT scores can have quite different ATE estimates.
 
\end{document}